# Can Artificial Intelligence Generate Quality Research Topics Reflecting Patients' Concerns?

Jiyeong Kim, PhD, MPH[1], Michael L. Chen, BS[1], Shawheen J. Rezaei, MPhil[1], Mariana Ramirez-Posada, MD[2], Jennifer L. Caswell-Jin, MD[3], Allison W. Kurian, MD, MSc[3,4], Fauzia Riaz, MD, MHS[3], Kavita Y. Sarin, MD, PhD[2], Jean Y. Tang, MD, PhD[2], Steven M. Asch, MD, MPH[1,5], Eleni Linos, MD, DrPH, MPH[1,2]

[1] Center for Digital Health, Stanford School of Medicine, Stanford, CA, USA
[2] Department of Dermatology, Stanford School of Medicine, Stanford, CA, USA
[3] Division of Oncology, Department of Medicine, Stanford School of Medicine, Stanford, CA, USA
[4] Department of Epidemiology and Population Health, Stanford School of Medicine, Stanford, CA, USA
[5] Division of Primary Care and Population Health, Stanford School of Medicine, Stanford, CA, USA

**Abstract**

Patient-centered research is increasingly important in narrowing the gap between research and patient care, yet incorporating patient perspectives into health research has been inconsistent. We created a case-study around an automated framework leveraging innovative natural language processing (NLP) and artificial intelligence (AI) with a large database of patient portal messages to generate research ideas that prioritize important patient issues. We further quantified the quality of AI-generated research topics. To define patients' clinical concerns, we analyzed 614,464 patient messages from 25,549 individuals with breast or skin cancer obtained from a large academic hospital (Stanford Health Care, 2013-2024), constructing a 2-staged unsupervised NLP topic model. Then, we generated research topics to resolve the defined issues using a widely used AI (ChatGPT-4o, OpenAI Inc., April 2024 version) with prompt-engineering strategies. We guided AI to perform multi-level tasks: 1) knowledge interpretation and summarization (e.g., interpreting and summarizing the NLP-defined topics), 2) knowledge generation (e.g., generating research ideas corresponding to patients' issues), 3) self-reflection and correction (e.g., ensuring and revising the research ideas after searching for scientific articles), and 4) self-reassurance (e.g., confirming and finalizing the research ideas). Six highly experienced breast oncologists and dermatologists assessed significance and novelty of AI-generated research topics using a 5-point Likert scale (1-exceptional; 5-poor). We calculated the mean ($m_{S\ [Significance]}$ and $m_{N\ [Novelty]}$) and standard deviation ($SD_{S\ [Significance]}$ and $SD_{N\ [Novelty]}$) for each topic. The overall average scores were $m_S=3.00$ [$SD_S=0.50$] and $m_N=3.29$ [$SD_N=0.74$] for breast cancer and $m_S=2.67$ [$SD_S=0.45$] and $m_N=3.09$ [$SD_N=0.68$] for skin cancer. One-third of the AI-suggested research topics were highly significant and novel when both scores were lower than the average. Notably, two-thirds of the AI-suggested topics were novel in both cancers. Our findings demonstrate that AI-generated research topics reflecting patient perspectives via large-volume patient messages can meaningfully guide future directions in patient-centered health research.

**Introduction**

Patient-centered research is increasingly important in narrowing the gap between research and patient care to facilitate clinically meaningful outcomes.[1] In the past decade, efforts to integrate patient perspectives into health research have informed treatment suggestions and regulatory guidelines. For instance, the Patient-Centered Outcomes Research Institute (PCORI) has actively supported patient engagement in observational health studies and advocated for incorporating patient-reported outcomes in clinical trials.[2,3] However, there are barriers to capturing patient perspectives through traditional health research methods, particularly due to the high resource burden (e.g., time, money, and energy) without a guarantee for productive patient engagement. The incorporation of patient perspectives into health research has consequently been inconsistent.[4,5]

Over the past decade, direct patient portal messaging in electronic health record (EHR) systems has become a mainstay of communication for patients to share their clinical questions and concerns with their clinicians.[6] The use of EHR-based communications has increased significantly during the COVID-19 pandemic, with the volume of secure patient messages more than doubling from 2020 to 2024.[7,8] While collecting patients' voices in volume is challenging and time- and resource-consuming, patient portal messages serve as a relatively unexplored resource for quickly identifying patients' clinical concerns and questions in real time.

Natural language processing (NLP) is a probability-based language model that can extract key information from large quantities of text.[9] With recent advances in artificial intelligence (AI), AI-based NLP has been rapidly adopted in research to analyze large volumes of patient- or public-generated text data, such as social media or patient forum data.[10] Large language models (LLMs) have shown impressive performance in a variety of research-related tasks, including providing article reviews to peer reviewers and aiding in patient selection for clinical trials.[11,12]

Using a large database of direct portal messages from patients with cancer, we propose an automated framework that uses AI-enabled NLP methods to prioritize the most important issues that patients discuss with their clinicians and generate patient-centered research topics from these concerns. We further validated the patient-centered research topics with domain experts. The lessons learned in this pilot study exemplify how AI-enhanced NLP can prioritize patient concerns to inform patient-centered research, patient counseling, and quality improvement opportunities.

**Methods**

*Data source and study design*

We obtained deidentified patient portal messages of individuals with breast or skin cancer, including melanoma, basal cell carcinoma, or squamous cell carcinoma, defined using ICD-10 codes from a large academic hospital (Stanford Health Care) and 22 affiliated centers in California (07/2013-04/2024). We only included messages labeled as a Patient Medical Advice Request (PMAR) that were routed to oncology from individuals with breast cancer or to dermatology from those with skin cancer. The Institutional Review Board at Stanford University approved this study.

*2-Staged topic modeling*

To identify patients' clinical concerns, we analyzed the secure messages, constructing a 2-staged unsupervised NLP topic model leveraging Bidirectional Encoder Representations from Transformers (BERT) and Balanced Iterative Reducing and Clustering using Hierarchies (BIRCH) techniques to extract the essential topics.[13] First, preprocessed messages were converted into sentences using pre-calculated embedding (all-miniLM-L6-v2).[14] To simplify the dimensionality, we applied Uniform Mapping and Approximation and Projection and

ConvectVectorizer to remove infrequent words. Finally, we employed a zero-shot clustering to categorize similar topics by cosine similarity score.

To further refine the clustered topics, we constructed a new BERTopic model, applying the BIRCH algorithm, which can effectively and efficiently manage large data.[15] We implemented Principal Component Analysis and incremental fitting techniques for additional data process efficiency.[16] Excluding administrative issues (e.g., scheduling/rescheduling appointments), we obtained the top five clinical concerns for breast and skin cancer groups. Full topic lists are in Table S1.

*AI to generate research topics*

With the defined clinical concerns, we generated research topics to help resolve the issues raised using a widely used LLM (ChatGPT-4o, OpenAI Inc., April 2024 version). To enhance the LLM's capacity, we applied prompt-engineering strategies consisting of multiple techniques to provide background context:[17,18] role prompting ("Dr. GPT, a professional oncologist in the hospital"), directive commanding ("summarize~," "suggest~," "search~,"), expertise emulation ("I myself am an oncologist"), and zero-shot chain of thought ("take time to think deeply and step-by-step").

Following prompt engineering, we guided the LLM to perform multi-level tasks. First, we directed it to interpret the NLP-defined topics using representative keywords regarding patients' clinical issues (knowledge interpretation and summarization). Second, we guided the LLM to generate research ideas corresponding to those patients' issues (knowledge generation). Third, to ensure its novelty, we instructed it to search electronic databases for scientific articles (e.g., PubMed, Cochrane) and revise its initially suggested research topics based on the search to fill knowledge gaps (self-reflection and correction). Lastly, we instructed that the LLM finalize the significance and novelty of the suggested research topics (self-reassurance). Full prompts are in Supplementary Method 1.

*Evaluation of AI-generated research topics*

Six highly experienced domain experts (three breast oncologists, A.W.K., J.L.C., and F.R., and three dermatologists, J.Y.T., E.L., and K.Y.S) assessed the significance and novelty of AI-generated research topics. To score the significance and novelty, raters used a 5-point Likert scale (1-exceptional; 2-outstanding; 3-good; 4-fair; 5-poor), a simplified scoring system of the National Institutes of Health (NIH) grant review scale (1-exceptional; 9-poor). Each assessor had 10-30 years of clinical practice, extensive research experience, and familiarity with the NIH grant review process. The assessors were asked to rate the significance and novelty of the research topic based on the NIH grant scoring process. For breast or skin cancer, we calculated the mean (m) and standard deviation (SD) of three assessments for each topic to present the level of agreement among assessors using ensembling.[19]

For novelty, we conducted an additional literature search to confirm that AI-generated research topics were novel. Two researchers (S.J.R. and M.P.) independently constructed search terms based on suggested research titles. The third researcher (J.K.) compared them and resolved conflicted search terms to finalize them (Table S2). Using the developed search terms, two researchers independently searched Google Scholar by relevance. We stopped screening the topic without reviewing all the search results and concluded that the research topic had not been published in the following two cases: 1) one of the search terms stopped appearing in the title and main text consistently for two pages, or 2) the 10$^{th}$ page in Google Scholar was reached. Given that the purpose of the search was to assess the novelty of the AI-generated topics, we did not perform a systematic literature search for each topic. Lastly, we reviewed the articles referenced by the AI-generated text to ensure that hallucinated content was not in its explanation to justify knowledge gaps. To quantify the concordance of two novelty assessments, experts' scores, and literature search results, we assessed the correlation between these two novelty measures using Spearman's rank correlation test. A full list of AI-drafted research topics with assessments is provided in Table S3.

**Results**

*Study population characteristics*

Table 1 shows the demographic characteristics of included patients with cancer (N=25,549; n=10,665 for breast and n=14,884 for skin). Most participants were non-Hispanic (91.7%). For breast cancer, 61.1% were white, 23.9% were Asian, and 98.6% were female. For skin cancer, the majority were white (88.7%), while sex was balanced (51% male and 49% female).

**Table 1. Demographic characteristics of patients with cancer (breast or skin) in secure messaging of SHC 2013-2024**

|  | Total cancer Frequency (N=25,549) | Percentage (%) | Breast cancer Frequency (n=10,665) | Percentage (%) | Skin cancer Frequency (n=14,884) | Percentage (%) |
|---|---|---|---|---|---|---|
| **Race** | | | | | | |
| Asian | 2,981 | 11.7 | 2,544 | 23.9 | 437 | 2.9 |
| Black | 265 | 1.0 | 223 | 2.1 | 42 | 0.3 |
| Native American/ Pacific Islander | 178 | 0.7 | 126 | 1.2 | 52 | 0.3 |
| Other | 1,754 | 6.8 | 1,100 | 10.3 | 654 | 4.4 |
| White | 19,718 | 77.2 | 6,518 | 61.1 | 13,200 | 88.7 |
| Unknown | 653 | 2.6 | 154 | 1.4 | 499 | 3.4 |
| **Ethnicity** | | | | | | |
| Hispanic | 1,284 | 5.0 | 863 | 8.1 | 421 | 2.8 |
| Non-Hispanic | 23,431 | 91.7 | 9,623 | 90.2 | 13,808 | 92.8 |
| Unknown | 834 | 3.3 | 179 | 1.7 | 655 | 4.4 |
| **Sex** | | | | | | |
| Female | 17,819 | 69.7 | 10,519 | 98.6 | 7,300 | 49.0 |
| Male | 7,730 | 30.3 | 146 | 1.4 | 7,584 | 51.0 |
| **Marital Status**[*] | | | | | | |
| Married | 17,531 | 68.6 | 7,295 | 68.4 | 10,236 | 68.8 |
| Unmarried[†] | 7,703 | 30.2 | 3,314 | 31.1 | 4,389 | 29.5 |
| Unknown | 307 | 1.2 | 55 | 0.5 | 252 | 1.7 |

[*] Marital status: Missingness in Skin cancer (n=7), Breast cancer (n=1), and Total (n=8);
[†] Unmarried included single, widowed, divorced, life partner, separated, and other.

*Patient message characteristics*

We obtained a total of 44,984,615 unique message threads from patients (breast cancer, n=1,679,390; and skin disease, n=43,305,225) from 2013-2024. Of those, 14,672,401 (32.6%) messages were labeled as PMAR. In this study, we analyzed 474,194 PMARs from patients with breast cancer routed to Oncology and 140,270 PMARs from patients with skin cancer routed to Dermatology (Figure 1).

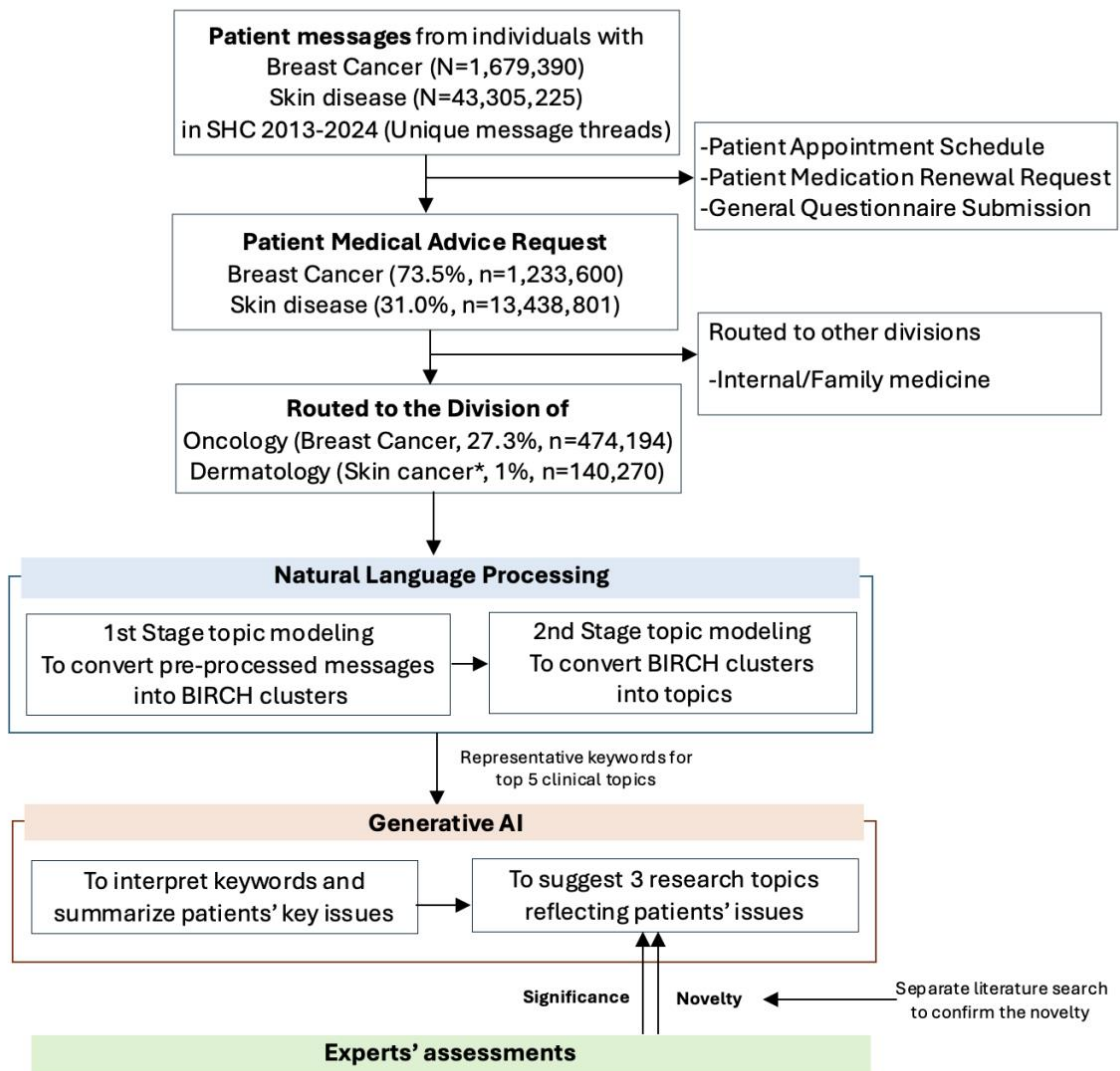

**Figure 1. Data source and study design**

* Skin cancer includes melanoma and other malignant neoplasms of the skin, including basal cell carcinoma and squamous cell carcinoma.

*Patients' clinical concerns*

Table 2 presents the primary clinical concerns of cancer patients interpreted by AI using NLP-generated keywords. In this study, we included the top five clinical concerns for each cancer, and AI identified three research topics for each clinical concern.

**Table 2. Primary clinical concerns* of patients with cancer (breast or skin) addressed through patient portal messages in 2013-2024**

| |
|---|
| **For patients with breast cancer** |
| **Topic 1. Keywords:** ['rash', 'itching', 'topical', 'allergic', 'lotion', 'redness'] |
| **†Concerns:** Skin-related issues (e.g., rashes and itching). These symptoms suggest possible allergic reactions or side effects from treatments, necessitating the use of topical treatments or lotions for alleviation. |
| **Topic 2. Keywords:** ['urine', 'urinalysis', 'urinating', 'bladder', 'peeing', 'cystitis'] |
| **Concerns:** Issues related to urinary function, including frequent urination, urinary tract infections (UTIs), and bladder discomfort. Concerns include symptoms of cystitis and the need for urinalysis to diagnose potential problems. These issues may be exacerbated by treatment side effects or infections. |
| **Topic 3. Keywords:** ['dentist', 'tooth', 'periodontist', 'oral', 'filling', 'endodontist'] |
| **Concerns:** Dental health issues (e.g., toothaches, dental procedures, and oral hygiene). They mention the need for visits to the dentist, periodontist, or endodontist for treatments (e.g., fillings and other procedures). These concerns highlight the importance of addressing oral health as part of their overall cancer care plan. |
| **Topic 4. Keywords:** ['genetic', 'geneticist', 'brca', 'testing', 'genomic'] |
| **Concerns:** Concerned about genetic testing, particularly BRCA gene mutations, and the implications for their diagnosis and treatment. They seek guidance from geneticists on the necessity and benefits of undergoing genetic testing and understanding the results, including BRCA2 mutations. |
| **Topic 5. Keywords:** ['liver', 'hepatic', 'ascites', 'biopsy', 'worried', 'ultrasound'] |
| **Concerns:** Liver-related issues (e.g., hepatic conditions and ascites). They are worried about results of biopsy and ultrasound. Patients are anxious about liver health and its implications on their overall cancer management. |
| **For patients with skin cancer** |
| **Topic 1. Keywords:** ['nasal', 'pimple', 'nostril', 'face', 'biopsy', 'surgery', 'skin', 'picture'] |
| **Concerns:** Issues about lesions or pimples on the nose and nostrils, which may resemble common skin issues but raise fears of malignancy. They often seek clarification on whether these lesions require a biopsy or surgery, particularly after sharing photos with their clinician for evaluation. |
| **Topic 2. Keywords:** ['mole', 'melanoma', 'concerned', 'removal', 'sure', 'skin', 'checked'] |
| **Concerns:** Issues about the potential for moles to develop into melanoma. They express anxiety over whether certain moles should be removed or further examined and seek reassurance or confirmation through appointments. The need for timely skin checks and possible mole removal is a significant focus of their concern. |
| **Topic 3. Keywords:** ['ear', 'earlobe', 'earrings', 'hearing', 'surgery', 'biopsy', 'cartilage', 'canal'] |
| **Concerns:** Issues about earlobe in relation to surgical interventions like biopsies and the impact on hearing. They also worried about complications involving the cartilage and ear canal, following surgeries or procedures. Additionally, they concerned about wearing earrings and the overall appearance of the ear post-surgery. |
| **‡Topic 4. Keywords:** ['effudex', 'treatment', 'week', 'biopsy'] |
| **Concerns:** Issues with the treatment process involving Efudex (Efudix/Effudex), focusing on its application duration, typically over several weeks, its effectiveness, and particularly, treatment impact following a biopsy. |
| **Topic 5. Keywords:** ['stitches', 'incision', 'sutures', 'surgery', 'wound', 'healing'] |

> **Concerns:** Issues about the management of their surgical wounds, related to stitches and sutures. Their messages reflect worries about incision care, the healing process, and potential complications with stitching after surgery. They seek guidance on proper wound care and expect timely advice on how to manage it.

\* The top five clinical issues were included in the study, excluding issues that required administrative support, including scheduling (e.g., appointments for radiation, chemotherapy, or infusion), authorization (e.g., billing or payment), and paperwork (e.g., Family and Medical Leave Act, FMLA form). AI model generated three research topics for each clinical concern;
† Concerns: LLM's interpretation using keywords extracted by the NLP model from patient messages;
‡ Keywords represent the most important and frequent issues of patient messages, which captured that the chemotherapy medication (5-FU; brand name [Efudex]) is commonly misspelled in patient messages, including Effudex or Efudix.

*Assessment of the quality of AI-generated research topics*

The overall significance score was lower than the novelty score in both breast ($m_S$=3.00 [$SD_S$=0.50] in Significance; $m_N$=3.29 [$SD_N$=0.74] in Novelty, 1-exceptional, 5-poor) and skin cancer ($m_S$=2.67 [$SD_S$=0.45] in Significance; $m_N$=3.09 [$SD_N$=0.68] in Novelty).

The most significant and novel research topics, where both scores were lower than the average scores ($m_S$=3.00 [$SD_S$=0.50]; $m_N$=3.29 [$SD_N$=0.74]), in breast cancer include 1) Interdisciplinary Approach to Managing Dental Health in Breast Cancer Care ($m_S$=2.33 [$SD_S$=1.15]; $m_N$=2.33 [$SD_N$=0.58]), 2) Evaluating the Efficacy of Hepatoprotective Agents in Preventing Liver Damage During Breast Cancer Treatment ($m_S$=2.33 [$SD_S$=1.15]; $m_N$=3.00 [$SD_N$=1.00]), 3) Longitudinal Study on the Impact of Genomic Testing on Treatment Outcomes ($m_S$=2.67 [$SD_S$=1.15]; $m_N$=3.00 [$SD_N$=1.00]), 4) Development and Testing of a Specialized Skin Care Regimen for Breast Cancer Patients ($m_S$=2.67 [$SD_S$=1.15]; $m_N$=3.00 [$SD_N$=1.00]) (Table 3).

For skin cancer, when the average scores were $m_S$=2.67 [$SD_S$=0.45] and $m_N$=3.09 [$SD_N$=0.68], both significant and novel topics included 1) Development and Evaluation of a Patient-Centered Digital Tool for Post-Surgical Wound Care ($m_S$=2.33 [$SD_S$=1.15]; $m_N$=2.33 [$SD_N$=0.58]), 2) Impact of Patient Education on Efudex Treatment Adherence and Outcomes ($m_S$=1.67 [$SD_S$=0.58]; $m_N$=2.33 [$SD_N$=1.53]), 3) Longitudinal Study on Patient Anxiety and Decision-Making in Mole Surveillance and Removal ($m_S$=2.33 [$SD_S$=1.53]; $m_N$=2.67 [$SD_N$=1.53]).

Approximately two-thirds of AI-generated research topics were found to be new, including breast cancer (n=10/15 topics) and skin cancer (n=11/15 topics). The experts' high novelty scores appeared to be positively correlated with no existing literature, yet not conclusive (ρ=0.28, p=0.13).

**Table 3. High-quality AI-generated research topics reflecting patients' concerns**

| | AI-generated research topics | Significance $m_S$ ($SD_S$)* | Novelty $m_N$ ($SD_N$)* |
|---|---|---|---|
| | **For patients with breast cancer** | | |
| 1 | Interdisciplinary Approach to Managing Dental Health in Breast Cancer Care | 2.33 (1.15) | 2.33 (0.58) |
| 2 | Evaluating the Efficacy of Hepatoprotective Agents in Preventing Liver Damage During Breast Cancer Treatment | 2.33 (1.53) | 3.00 (1.00) |
| 3 | Longitudinal Study on the Impact of Genomic Testing on Treatment Outcomes | 2.67 (1.15) | 3.00 (1.00) |
| 4 | Development and Testing of a Specialized Skin Care Regimen for Breast Cancer Patients | 2.67 (1.15) | 3.00 (1.00) |
| 5 | Efficacy of Preventive Dental Care Protocols for Breast Cancer Patients | 3.00 (1.00) | 2.33 (0.58) |
| | **For patients with skin cancer** | | |
| 1 | Development and Evaluation of a Patient-Centered Digital Tool for Post-Surgical Wound Care | 2.00 (0.00) | 1.67 (0.58) |
| 2 | Impact of Patient Education on Efudex Treatment Adherence and Outcomes | 1.67 (0.58) | 2.33 (1.53) |
| 3 | Longitudinal Study on Patient Anxiety and Decision-Making in Mole Surveillance and Removal | 2.33 (1.53) | 2.67 (1.53) |
| 4 | Impact of Ear and Earlobe Reconstruction on Hearing Post Skin Cancer Surgery | 2.33 (0.58) | 3.00 (1.73) |
| 5 | Impact of Suture Materials on Scar Formation in Skin Cancer Patients | 2.67 (1.15) | 2.67 (0.58) |
| 6 | Investigation of Cartilage-Sparing Techniques in Skin Cancer Surgery of the Ear | 2.67 (1.15) | 2.67 (1.15) |

\* Mean (m) and SD scores were computed based on the scores of three experts; Among 30 AI-generated research topics (15 in breast cancer and 15 in skin cancer), only high-quality topics were listed in this table in order of quality (from high to low) when both scores were better than the overall mean (A lower score means better quality). A full list of AI-generated research topics with experts' scores is available in Table S3.

**Discussion**

In this study, we assessed an LLM's capacity to generate patient-centered research topics and evaluated the significance and novelty of the AI-generated research topics. Approximately one-third of the AI-suggested research topics were considered highly significant and novel for both breast and skin cancer. Notably, two-thirds of the AI-suggested topics were novel in both cancer groups. Research topics generated for skin cancer patients were more significant and novel overall compared to those generated for breast cancer patients. The findings highlight that AI/NLP-based research question creation is a promising way to promote patient-centered research tailored specifically to patients' most pressing concerns.

Collecting patients' perspectives in volume is challenging due to the time-intensity of qualitative interviews and the resource barriers involved in conducting qualitative studies. Moreover, patient priorities continuously evolve, and the relative importance of different patient-reported concerns may not be clear during the conception of new research studies. As a result, there are significant barriers to designing and completing research studies that address the most important concerns facing patients. Our AI-enabled NLP pilot study provides a quantitative and repeatable way to identify the most important patient concerns, offering an opportunity to bridge patients' issues and health research. The AI-enabled NLP model allowed us to define and understand the top five clinical issues for patients with cancer through systematic analysis of 614,464 unique messages from 25,549 individuals in the past 10 years. AI-based research topic generation identified current knowledge gaps that were scientifically significant and novel research topics in patient-informed research.

We acknowledge limitations of our study. First, we generated 30 research questions for two medical specialty areas, and our findings may not be generalizable to other medical conditions or non-users of patient messaging. Second, we excluded patients' needs for administrative support (e.g., scheduling, medication refills, or insurance issues) and prioritized identifying clinical issues that were most relevant to patient-centered research outcomes. Separate research is warranted to identify patients' issues with health services to inform administrative support interventions, given their high volume of interest. Third, the domain experts were all from a single academic institution, which might have brought potential bias in scoring. However, assessors' research and clinical areas spanned diverse subfields. We suggest further rigorous evaluations with larger sample topics in various specialty areas to incorporate diverse assessors and patient viewpoints. Moreover, evaluating patients' perspectives on AI-generated research topics would be critical to discerning genuinely essential topics for them. Exploring ways to invite investigators and funding agencies to collaborate on these topics inspired by patients' priorities would also be beneficial.

In summary, our pilot study illustrates how an AI/NLP-based automated framework can systematically prioritize patient concerns to inform patient-centered research. Our findings that the AI-generated research topics were high quality and scientifically novel demonstrate that using patient perspectives via large-volume patient message data can meaningfully guide future directions in health research.

**Supplementary**

**Can Artificial Intelligence Generate Quality Research Topics Reflecting Patients' Concerns?**

**Table of Content**

**Table S1. A full list of key topics and keywords that NLP defined after analyzing patient portal messages from those with breast or skin cancer, Stanford Health Care 2013-2024**

| Skin cancer: key topics | Example keywords |
|---|---|
| 0_pharmacies_prescription_pharmacy_prescriptions | ['pharmacies', 'prescription', 'pharmacy', 'prescriptions', 'walgreen', 'walgreens', 'pharmacist', 'cvs', 'prescribed', 'medications'] |
| 1_photos_pictures_pic_photo | ['photos', 'pictures', 'pic', 'photo', 'pics', 'images', 'sending', 'send', 'image', 'message'] |
| 2_10am_15pm_30pm_15am | ['10am', '15pm', '30pm', '15am', '30am', '45am', '45pm', '1pm', 'scheduling', 'noon'] |
| 3_appointment_appointments_rescheduled_scheduled | ['appointment', 'appointments', 'rescheduled', 'scheduled', 'reschedule', 'scheduling', 'schedule', 'calendar', 'cancellation', 'cancel'] |
| 4_insurance_billing_medicare_billed | ['insurance', 'billing', 'medicare', 'billed', 'coverage', 'deductible', 'healthcare', 'payment', 'copay', 'provider'] |
| 5_surgery_surgeries_appointment_surgical | ['surgery', 'surgeries', 'appointment', 'surgical', 'scheduled', 'surgeon', 'schedule', 'scheduling', 'procedure', 'anesthesia'] |
| 6_nose_nasal_pimple_nostril | ['nose', 'nasal', 'pimple', 'nostril', 'face', 'biopsy', 'surgery', 'photo', 'skin', 'picture'] |
| 7_biopsy_pathology_biopsies_results | ['biopsy', 'pathology', 'biopsies', 'results', 'received', 'dr', 'review', 'procedure', 'reports', 'benign'] |
| 8_dermatologist_dermatologists_dermatology_dermatological | ['dermatologist', 'dermatologists', 'dermatology', 'dermatological', 'dermatitis', 'dermatologic', 'clinic', 'appointment', 'doctor', 'patients'] |
| 9_mole_moles_melanoma_concerned | ['mole', 'moles', 'melanoma', 'concerned', 'appointment', 'been', 'removal', 'sure', 'skin', 'checked'] |
| 10_ear_earlobe_ears_earrings | ['ear', 'earlobe', 'ears', 'earrings', 'hearing', 'surgery', 'lobe', 'biopsy', 'cartilage', 'canal'] |
| 11_efudix_efudex_effudex_treatment | ['efudix', 'efudex', 'effudex', 'treatment', 'treat', 'treated', 'treating', 'weeks', 'week', 'biopsy'] |
| 12_stitches_incision_stitch_sutures | ['stitches', 'incision', 'stitch', 'sutures', 'stitching', 'surgery', 'wound', 'healing', 'suture', 'sutured'] |
| **Breast cancer: Key topics** | **Example keywords** |
| 0_authorization_submitted_billing_payment | ['authorization', 'submitted', 'billing', 'payment', 'received', 'authorized', 'provider', 'approval', 'approved', 'billed'] |
| 1_rash_itching_rashes_itchy | ['rash', 'itching', 'rashes', 'itchy', 'itchiness', 'itch', 'topical', 'allergic', 'lotion', 'redness'] |
| 2_radiotherapy_radiation_appointment_oncology | ['radiotherapy', 'radiation', 'appointment', 'oncology', 'scheduled', 'appt', 'treatment', 'therapy', 'scheduling', 'oncologist'] |
| 3_resulted_result_testing_results | ['resulted', 'result', 'testing', 'results', 'received', 'tested', 'checking', 'test', 'question', 'ordered'] |
| 4_appointment_scheduled_appt_question | ['appointment', 'scheduled', 'appt', 'question', 'video', 'received', 'visit', 'talk', 'videos', '30am'] |
| 5_urine_urinalysis_urinating_urination | ['urine', 'urinalysis', 'urinating', 'urination', 'urinary', 'utis', 'urinate', 'bladder', 'peeing', 'cystitis'] |
| 6_vaccination_vaccine_vaccinations_vaccinate | ['vaccination', 'vaccine', 'vaccinations', 'vaccinate', 'vaccines', 'appointment', 'shot', 'scheduled', 'covid', 'reschedule'] |
| 7_appointment_scheduled_reschedule_scheduling | ['appointment', 'scheduled', 'reschedule', 'scheduling', 'surgery', 'procedure', 'waiting', 'surgical', 'week', 'schedule'] |
| 8_pleasanton_norcal_emeryville_oakland | ['pleasanton', 'norcal', 'emeryville', 'oakland', 'near', 'available', 'appointment', 'alto', 'redwood', 'need'] |
| 9_disability_fmla_submitted_paperwork | ['disability', 'fmla', 'submitted', 'paperwork', 'filed', 'extended', 'requesting', 'regarding', 'need', 'absence'] |
| 10_appointment_scheduled_sledge_reschedule | ['appointment', 'scheduled', 'sledge', 'reschedule', 'patient', 'received', 'appt', 'mentioned', 'regarding', 'asked'] |
| 11_chemo_appointment_chemotherapy_scheduled | ['chemo', 'appointment', 'chemotherapy', 'scheduled', 'reschedule', 'oncology', 'cancer', 'scheduling', 'appt', 'treatment'] |
| 12_dentist_tooth_dental_teeth | ['dentist', 'tooth', 'dental', 'teeth', 'periodontist', 'oral', 'mouth', 'filling', 'endodontist', 'procedure'] |
| 13_mammogram_mammography_appointment_mamogram | ['mammogram', 'mammography', 'appointment', 'mamogram', 'scheduled', 'breast', 'appt', 'scheduling', 'screening', 'ordered'] |
| 14_infusion_appointment_scheduled_scheduling | ['infusion', 'appointment', 'scheduled', 'scheduling', 'appt', 'schedule', 'week', 'appts', 'tomorrow', 'need'] |
| 15_pharmacy_prescription_prescribed_pharmacist | ['pharmacy', 'prescription', 'prescribed', 'pharmacist', 'walgreens', 'contacted', 'medication', 'refill', 'received', 'refills'] |
| 16_photograph_picture_pic_uploaded | ['photograph', 'picture', 'pic', 'uploaded', 'updated', 'image', 'photos', 'attached', 'yesterday', 'camera'] |
| 17_genetic_geneticist_genetics_brca | ['genetic', 'geneticist', 'genetics', 'brca', 'testing', 'diagnosed', 'genomic', 'tested', 'gene', 'brca2'] |
| 18_covid_covid19_tested_scheduled | ['covid', 'covid19', 'tested', 'scheduled', 'swab', 'testing', 'sick', 'test', 'symptom', 'quarantine'] |
| 19_appointment_scheduled_patient | ['appointment', 'scheduled', 'patient', 'referral', 'reschedule', 'question', 'appt', 'regarding', 'scheduling'] |
| 20_treated_treatment_patient_therapy | ['treated', 'treatment', 'patient', 'therapy', 'discussed', 'plan', 'question', 'regarding', 'treat', 'understanding'] |
| 21_suggestion_consideration_appreciate_response | ['suggestion', 'consideration', 'appreciate', 'response', 'thank', 'apologies', 'attention', 'appreciation', 'situation', 'accept'] |
| 22_appointment_radiology_radiologist_scheduled | ['appointment', 'radiology', 'radiologist', 'scheduled', 'reschedule', 'scheduling', 'referral', 'appt', 'contacted', 'imaging'] |
| 23_liver_hepatology_hepatic_ascites | ['liver', 'hepatology', 'hepatic', 'ascites', 'biopsy', 'question', 'concerned', 'worried', 'showed', 'ultrasound'] |

*Pink colored cells are selected clinical topics for this study.

**Supplementary Method 1. Engineered prompts used to interpret the key topics from NLP and draft research topics tailored to such topics**

**1) Prompts for messages from patients with breast cancer**

**Researcher** *[provided context using multiple techniques, including role prompting, directive commanding, expertise emulation, and zero-shot chain of thought]*:
As Dr. GPT, a professional oncologist in the hospital, one of your roles is responding to the patients' messages through the patient portal. Patients with breast cancer ask health questions, reporting some symptoms or using lab results. We performed topic modeling using patient messages, and obtained representative keywords for each topic. I will ask for your help interpreting the topic using the keywords.

I myself am an oncologist in the hospital. I will use your response to better understand patients' needs from clinicians. Your final response should be concise but concrete and comprehensive in a professional manner using your specialty knowledge and experience in oncology, especially in breast cancer. For each question, you should take time to think deeply and step-by-step to be sure to offer the right answer.

To begin, please confirm that you understand your role and express your preparedness to help me before providing any information.

**AI:** (answers)

**Researcher** *[Task: knowledge interpretation and summarization]*:
Using the Topic_label and Keywords, can you summarize the primary issues that patients with breast cancer are experiencing in less than three sentences?

**AI:** (answers)

**Researcher:**
Thank you!

**AI:** (answers)

**Researcher** *[Task: knowledge generation]*:
Can you also suggest optimal research projects that can help resolve patients' issues yet have not been conducted or published as an article? Please limit the number of research projects to 3.

**AI:** (answers)

**Researcher** *[Task: self-reflection and self-correction]*:
Are you sure that these have not been conducted by other researchers or published in journals? Can you search at least 4 web-based databases, including PubMed, Cochrane, Web of Science, and Embase, to make sure these research projects have not yet been published in the past 15 years?

**Researcher** *[Task: self-reassurance]*:
If you can confirm that, could you clarify the objectives, significance, and novelty of each project?

**Researcher:**
Great job, thank you!

**2) Prompts for messages from patients with skin cancer**

**Researcher:**
As Dr. GPT, a professional dermatologist in the hospital, one of your roles is responding to the patients' messages through the patient portal. Patients ask health questions, reporting some symptoms or using lab results. We performed topic modeling using patient messages, and obtained representative keywords for each topic. I will ask for your help interpreting the topic using the keywords.

I myself am a dermatologist in the hospital. I will use your response to better understand patients' needs from clinicians. Your final response should be concise but concrete and comprehensive in a professional manner using your specialty knowledge and experience in dermatology. For each question, you should take time to think deeply and step-by-step to be sure to offer the right answer.

To begin, please confirm that you understand your role and express your preparedness to help me before providing any information.

**AI:** (answers)

**Researcher** *[Task: knowledge interpretation and summarization]*:
Using the Topic_label and Keywords, can you summarize the primary issues that patients with skin cancer are experiencing in less than three sentences?

**AI:** (answers)

**Researcher:**
Thank you!

**AI:** (answers)

**Researcher** *[Task: knowledge generation]*:
Can you also suggest optimal research projects that can help resolve patients' issues yet have not been conducted or published as an article? Please limit the number of research projects to 3.

**AI:** (answers)

**Researcher** *[Task: self-reflection and self-correction]*:

Are you sure that these research projects have not been conducted by other researchers or published in journals? Can you search at least four web-based databases, including PubMed, Cochrane, Web of Science, and Embase, to ensure that these research projects have not been published in the past 15 years?

**AI:** (answers)

**Researcher** *[Task: self-reassurance]*:
If you can confirm that, could you clarify the objectives, significance, and novelty of each project?

**AI:** (answers)

**Researcher:**
Great job, thank you!

**AI:** (answers)

**Table S2. Literature search results and rationales for final decision**

| Topics* | Rater A | Rater B | Rater C† (Final) | Rationales for final decision |
|---|---|---|---|---|
| BC_1_2 | Maybe | Maybe | No | Agreed with the given rationale from AI ("While there is existing research on skin toxicities related to specific cancer medications, a detailed investigation that combines data on allergic reactions to both cancer medications and commonly used skin care products is not comprehensively covered"). One study assessed the toxicity of aloe vera gel in the management of radiation induced skin reaction in breast cancer in 2006, which may need to be updated. |
| BC_2_1 | Maybe | Yes | Yes | e.g., https://www.sciencedirect.com/science/article/pii/S0085253815596993 |
| BC_4_2 | Maybe | Yes | No | Theoretically designed interventions can be effective in helping women understand their cancer risk and appropriate risk assessment options. However, the previous study used theoretical approach and it was 17 years ago, hence a new study would be worthwhile. |
| BC_4_3 | Maybe | No | No | The key part of the suggested topic is a longitudinal study to see the long-term impact of genomic testing on treatment outcomes. Existing studies did not assess the long-term outcomes. |
| BC_5_1 | Maybe | No | No | Although there were studies that assessed the hepatotoxicity of radio therapy or doxorubicine among breast cancer, systematic investigations of hepatotoxicity of various chemotherapies can be done. |
| BC_5_3 | No | Yes | Yes | e.g., https://journals.sagepub.com/doi/full/10.1177/10781552241268778 |
| SK_1_2 | Maybe | No | No | Essentially, one study was done more than 16 years ago for one non-surgical approach on 4 patients. Hence, a new study would be worth pursuing, and we would consider this as novel. |
| SK_1_3 | Yes | No | Yes | e.g.,https://www.sciencedirect.com/science/article/pii/S1748681512001076?via%3Dihub |
| SK_2_1 | No | Maybe | No | Not the same study was done what incorporates all three components of AI-based algorithms: 1) to assess changes in moles over time, 2) provide risk assessments, and 3) with options for direct communication with dermatologists for high-risk cases. |
| SK_2_3 | Maybe | Maybe | Yes | e.g., https://researchspace.auckland.ac.nz/handle/2292/59230 |
| SK_5_1 | No | Maybe | Yes | e.g., https://academic.oup.com/ced/article/46/8/1400/6598700 |
| SK_5_2 | Yes | Maybe | No | Although digital tools for post-surgical wound care exist (e.g., https://digital.lib.washington.edu/researchworks/items/d47ac863-c07d-45e6-8546-3ab0f42872bf). None tailored specifically to the needs of skin cancer patients. |

* BC_1_2: Breast cancer topics 1, research topic 2.
† Yes (at least one article with almost the same research question exists) and No (no same article exists). The key determinative factor was the question to ourselves: Can we justify conducting AI-suggested research topic and envision adding new knowledge to the field? Yes vs No.

**Table S3. A full list of scores of domain experts for AI-generated research topics**

| |
|---|
| **Breast cancer** |
| **Topic 1.** rash_itching_rashes_itchy |
| **Primary issue**: Patients with breast cancer are experiencing skin-related issues, primarily rashes and itching. These symptoms suggest possible allergic reactions or side effects from treatments, necessitating the use of topical treatments or lotions to alleviate redness and itchiness. Addressing these concerns with appropriate skin care recommendations and evaluating for potential allergic reactions is crucial. |
| **1) Comparative Study on the Efficacy of Topical Treatments for Rash and Itching in Breast Cancer Patients** <br><br> Objective: <br> To evaluate and compare the effectiveness of various topical treatments (e.g., corticosteroids, antihistamines, moisturizing lotions) in alleviating rashes and itching caused by breast cancer treatments. <br><br> **Significance**: <br> This study aims to identify the most effective topical treatments for managing skin-related side effects in breast cancer patients. By providing clear evidence on which treatments work best, clinicians can offer targeted and effective solutions, improving patients' quality of life and adherence to cancer treatments. <br><br> **Novelty**: <br> While individual studies on specific topical treatments exist, a comprehensive comparative analysis across a wide range of treatments for rashes and itching due to various breast cancer treatments (not limited to radiation therapy) is lacking. This research will fill that gap by providing a holistic understanding of the efficacy of different topical treatments. |
| Significance (experts' scores): $m_S$=3.00 ($SD_S$=1.00) |
| Novelty (experts' scores): $m_N$=3.67 ($SD_N$= 1.15) |
| Novelty (literature search): Y (Yes, there was the same research that was already done) |
| Search terms: (Efficacy) AND (Topical Treatments) AND (Rash OR Itching) AND (Breast Cancer) |
| **2) Investigation of Allergic Reactions to Breast Cancer Medications and Skin Care Products** <br><br> Objective: <br> To examine the prevalence and triggers of allergic reactions to chemotherapy drugs, hormonal therapies, and commonly used skin care products in breast cancer patients. <br><br> **Significance**: <br> Understanding the specific allergens and irritants can lead to better management and prevention of skin reactions. This knowledge will allow for personalized skin care recommendations and adjustments to treatment regimens, minimizing discomfort and enhancing patient outcomes. <br><br> **Novelty**: <br> While there is existing research on skin toxicities related to specific cancer medications, a detailed investigation that combines data on allergic reactions to both cancer medications and |
| commonly used skin care products is not comprehensively covered. This project will provide a more integrated approach to understanding and managing allergic reactions in breast cancer patients. |
| Significance (experts' scores): $m_S$=3.67 ($SD_S$=1.53) |
| Novelty (experts' scores): $m_N$=2.67 ($SD_N$=0.58) |
| Novelty (literature search): N (No, there was no same research) |
| Search terms: (Allergic reactions) AND (breast cancer medications) AND (skin care products) |
| **3) Development and Testing of a Specialized Skin Care Regimen for Breast Cancer Patients**<br><br>Objective:<br>To create and test a comprehensive skin care regimen designed specifically for breast cancer patients undergoing treatment, aimed at preventing and managing rashes and itching.<br><br>**Significance**:<br>A specialized regimen can proactively address skin issues, improving treatment adherence and patient comfort. This can lead to better overall treatment outcomes and enhance the quality of life for breast cancer patients.<br><br>**Novelty**:<br>Although there are studies on specific skin care treatments for radiation dermatitis, a holistic regimen tailored to the unique needs of breast cancer patients undergoing various treatments (chemotherapy, hormonal therapy, etc.) has not been extensively explored. This research will develop and validate a comprehensive skin care approach, setting a new standard for patient care in oncology. |
| Significance (experts' scores): $m_S$=2.67 ($SD_S$=1.15) |
| Novelty (experts' scores): $m_N$=3.00 ($SD_N$=1.00) |
| Novelty (literature search): N (No, there was no same research) |
| Search terms: (Skin care regimen) AND (Breast Cancer) |
| **Topic 2**. urine_urinalysis_urinating_urination |
| **Primary issue**: Patients with breast cancer are reporting issues related to urinary function, including frequent urination, urinary tract infections (UTIs), and bladder discomfort. Concerns include symptoms of cystitis and the need for urinalysis to diagnose potential problems. These issues may be exacerbated by treatment side effects or infections. |
| **1) Impact of Chemotherapy on Urinary Function in Breast Cancer Patients**<br><br>Objective:<br>To investigate the prevalence, severity, and types of urinary issues (such as frequent urination, UTIs, and cystitis) among breast cancer patients undergoing chemotherapy.<br>To identify any correlations between specific chemotherapy regimens and urinary complications.<br><br>**Significance**:<br>Understanding how chemotherapy affects urinary function can help healthcare providers anticipate and manage these side effects more effectively.<br>Improved management of urinary issues can enhance the overall quality of life for breast cancer patients during and after treatment.<br><br>**Novelty**: |
| While there is some research on chemotherapy side effects, few studies have focused specifically on urinary function in breast cancer patients. This project would address a significant gap in the literature, providing new insights into a relatively underexplored area. |
| Significance (experts' scores): $m_S$=4.00 ($SD_S$=1.00) |
| Novelty (experts' scores): $m_N$=3.00 ($SD_N$=1.00) |
| Novelty (literature search): Y (Yes, there was the same research that was already done) |
| Search terms: (Chemotherapy) AND (Urinary Function) AND (Breast Cancer) |

**2) Preventive Measures for Urinary Tract Infections in Breast Cancer Patients**

Objective:
To evaluate the effectiveness of various preventive measures, such as probiotics, hydration protocols, and antibiotic prophylaxis, in reducing the incidence of UTIs in breast cancer patients.
To determine the best practices for preventing UTIs in this patient population.

**Significance**:
UTIs can cause significant discomfort and lead to serious complications if left untreated. Preventing these infections can improve patient outcomes and reduce healthcare costs.
By identifying effective preventive strategies, clinicians can offer better care tailored to the needs of breast cancer patients.

**Novelty**:
This research would specifically target breast cancer patients, a group that may have unique risk factors and needs. Most existing studies on UTI prevention are general and do not focus on this specific population.

| |
|---|
| Significance (experts' scores): $m_S$=3.00 ($SD_S$=1.00) |
| Novelty (experts' scores): $m_N$=3.33 ($SD_N$=0.58) |
| Novelty (literature search): N (No, there was no same research) |
| Search terms: (Preventive Measures) AND (Urinary tract infections) AND (Breast Cancer) |

**3) Bladder Health Management in Breast Cancer Survivors**

Objective:
To develop and test a comprehensive bladder health management program that includes dietary modifications, pelvic floor exercises, and regular monitoring.
To assess the long-term effectiveness of the program in reducing chronic urinary issues among breast cancer survivors.

**Significance**:
Chronic urinary issues can significantly impact the quality of life for breast cancer survivors. A targeted management program can provide lasting benefits and support long-term survivorship care.
Implementing a comprehensive approach can address multiple aspects of bladder health, providing a holistic solution.

**Novelty**:
There is a lack of research focused on long-term bladder health management for breast cancer survivors. This project would be one of the first to propose and test a comprehensive program, filling a crucial gap in post-treatment care.

| |
|---|
| Significance (experts' scores): $m_S$=3.33 ($SD_S$=0.58) |
| Novelty (experts' scores): $m_N$=2.67 ($SD_N$=0.58) |
| Novelty (literature search): N (No, there was no same research) |
| Search terms: (Bladder health) AND (Breast Cancer Survivors) |
| **Topic 3**. dentist_tooth_dental_teeth |
| **Primary issue**: Patients with breast cancer are expressing concerns related to dental health, including issues such as toothaches, dental procedures, and oral hygiene. They mention the need for visits to the dentist, periodontist, or endodontist for treatments like fillings and other dental procedures. These concerns highlight the importance of addressing oral health as part of their overall cancer care plan. |
| **1) Impact of Chemotherapy on Oral Health in Breast Cancer Patients**<br><br>Objective:<br>To investigate the effects of chemotherapy on the oral health of breast cancer patients, focusing on issues such as tooth decay, gum disease, oral mucositis, and changes in oral microbiota.<br><br>**Significance**:<br>Chemotherapy is known to cause various oral complications, but specific impacts on breast cancer patients need targeted exploration. Understanding these effects can help in developing better preventive and treatment strategies, thereby improving the quality of life for these patients.<br><br>**Novelty**:<br>While there is considerable information on general oral health impacts due to chemotherapy, detailed research specifically targeting the chemotherapeutic regimens used in breast cancer and their long-term oral health effects is scarce. This project aims to fill this gap by providing targeted insights for this patient group, enhancing both clinical management and patient education. |
| Significance (experts' scores): $m_S$=2.67 ($SD_S$=1.53) |
| Novelty (experts' scores): $m_N$=3.67 ($SD_N$=1.15) |
| Novelty (literature search): Y (Yes, there was the same research that was already done) |
| Search terms: (Chemotherapy) AND (Oral health) AND (Breast cancer) |
| **2) Efficacy of Preventive Dental Care Protocols for Breast Cancer Patients**<br><br>Objective:<br>To evaluate the effectiveness of tailored preventive dental care protocols, including regular dental check-ups, specialized oral hygiene practices, and prophylactic treatments, in reducing dental issues in breast cancer patients.<br><br>**Significance**:<br>Implementing effective preventive measures can help mitigate oral health problems, which are common during cancer treatment and can lead to severe complications if not managed properly. Improved dental health can result in fewer interruptions to cancer treatment and better overall health outcomes.<br><br>**Novelty**:<br>Existing guidelines and reviews address preventive dental care for cancer patients in general, but specific protocols designed and evaluated exclusively for breast cancer patients are not well- |
| documented. This research would pioneer the development of such protocols, addressing the unique oral health challenges faced by breast cancer patients. |
| Significance (experts' scores): $m_S$=3.00 ($SD_S$=1.00) |
| Novelty (experts' scores): $m_N$=2.33 ($SD_N$=0.58) |
| Novelty (literature search): N (No, there was no same research) |
| Search terms: (Efficacy) AND (Preventive) AND (Dental Care) AND (Breast Cancer) |
| **3) Interdisciplinary Approach to Managing Dental Health in Breast Cancer Care**<br><br>Objective:<br>To develop and assess an interdisciplinary care model that integrates dental and oncology care for breast cancer patients, facilitating communication and coordination between dentists and oncologists.<br><br>**Significance:**<br>A coordinated approach can enhance the management of oral health issues, ensuring that dental complications are promptly addressed and do not interfere with cancer treatment. This model aims to improve patient outcomes through better communication and integrated care.<br><br>**Novelty:**<br>While the integration of dental and oncology care has been recommended, formalized studies and models demonstrating the efficacy and outcomes of such interdisciplinary approaches specifically for breast cancer patients are limited. This project would be the first to develop and assess a comprehensive interdisciplinary care model, setting a new standard for holistic patient management and care coordination. |
| Significance (experts' scores): $m_S$=2.33 ($SD_S$=1.15) |
| Novelty (experts' scores): $m_N$=2.33 ($SD_N$=0.58) |
| Novelty (literature search): N (No, there was no same research) |
| Search terms: (Interdisciplinary Approach) AND (Dental health) AND (Breast Cancer) |
| **Topic 4**. genetic_geneticist_genetics_brca |
| **Primary issue**: Patients with breast cancer are concerned about genetic testing, particularly BRCA gene mutations, and the implications for their diagnosis and treatment. They seek guidance from geneticists on the necessity and benefits of undergoing genetic testing and understanding the results, including BRCA2 mutations. These concerns highlight the need for clear communication about the role of genetics in breast cancer management and potential preventative measures for themselves and their families. |
| **1) Study on the Psychological Impact of Genetic Testing on Breast Cancer Patients**<br><br>Objective:<br>To assess the emotional and mental health effects of genetic testing for BRCA mutations among breast cancer patients.<br><br>**Significance**:<br>This study aims to develop supportive interventions to mitigate the psychological stress associated with genetic testing, enhancing overall patient well-being.<br><br>**Novelty**: |
| It provides a comprehensive analysis of the psychological aspects of genetic testing, which is under-researched compared to the clinical benefits. |
| Significance (experts' scores): $m_S$=3.33 ($SD_S$=2.08) |
| Novelty (experts' scores): $m_N$=5.00 ($SD_N$=0.00) |
| Novelty (literature search): Y (Yes, there was the same research that was already done) |
| Search terms: (Psychological Impact) AND (Genetic Testing) AND (Breast Cancer) |
| **2) Effectiveness of Genetic Counseling Interventions in Decision-Making**<br><br>Objective:<br>To evaluate the role of genetic counseling in aiding patients' understanding and decision-making regarding genetic testing and preventive measures.<br><br>**Significance:**<br>The findings could improve genetic counseling practices, ensuring patients make well-informed decisions based on clear understanding.<br><br>**Novelty:**<br>This research fills a gap by focusing on the decision-making process facilitated by genetic counseling, which has not been extensively studied. |
| Significance (experts' scores): $m_S$=2.67 ($SD_S$=1.53) |
| Novelty (experts' scores): $m_N$=4.33 ($SD_N$=1.15) |
| Novelty (literature search): N (No, there was no same research) |
| Search terms: (Genetic Counseling) AND (Decision-Making) AND (Breast cancer) |
| **3) Longitudinal Study on the Impact of Genomic Testing on Treatment Outcomes**<br><br>Objective:<br>To investigate the long-term effects of genomic testing on treatment decisions and outcomes in breast cancer patients.<br><br>**Significance**:<br>The study could lead to more personalized and effective treatment plans, improving patient outcomes and quality of life.<br><br>**Novelty**:<br>Unlike existing research focused on short-term impacts, this study examines the enduring benefits and challenges of integrating genomic data into long-term treatment planning. |
| Significance (experts' scores): $m_S$=2.67 ($SD_S$=1.15) |
| Novelty (experts' scores): $m_N$=3.00 ($SD_N$=1.00) |
| Novelty (literature search): N (No, there was no same research) |
| Search terms: (Genomic Testing) AND (Treatment Outcomes) AND (Breast cancer) |
| **Topic 5**. liver_hepatology_hepatic_ascites |
| **Primary issue**: Patients with breast cancer are expressing concerns about liver-related issues, specifically hepatic conditions and ascites. They are worried about biopsy results and ultrasound findings, often seeking clarification and reassurance from their oncologist. These patients are anxious about liver health and its implications on their overall cancer management. |
| **1) Study on the Impact of Breast Cancer Treatments on Liver Function** |

Objective:
To systematically investigate the effects of various breast cancer treatments, including chemotherapy and targeted therapies, on liver function and health over time.

**Significance**:
Breast cancer treatments, particularly chemotherapy and targeted therapies, have known hepatotoxic effects that can impact liver function. Understanding these effects is crucial for managing side effects and improving the overall health and treatment outcomes of breast cancer patients. This research can lead to better monitoring protocols and potentially the development of strategies to mitigate liver damage during treatment.

**Novelty:**
While there is some research on the hepatotoxic effects of specific drugs, a comprehensive study focusing on the overall impact of breast cancer treatments on liver function has not been conducted. Most existing studies examine individual drugs or specific liver complications rather than taking a holistic approach to understand the cumulative impact of breast cancer treatments on liver health.

| | |
|---|---|
| Significance (experts' scores): $m_S$=2.67 ($SD_S$=1.53) | |
| Novelty (experts' scores): $m_N$=3.33 ($SD_N$=0.58) | |
| Novelty (literature search): N (No, there was no same research) | |
| Search terms: (Breast Cancer Treatments) AND (Liver Function) | |

**2) Development of Non-Invasive Diagnostic Tools for Early Detection of Hepatic Complications in Breast Cancer Patients**

Objective:
To develop and validate non-invasive diagnostic tools, such as advanced imaging techniques and biomarkers, for the early detection of hepatic complications in breast cancer patients undergoing treatment.

**Significance:**
Early detection of liver issues can significantly enhance the management of breast cancer treatment, allowing for timely interventions that reduce morbidity and improve patient outcomes. Non-invasive diagnostic tools will provide a safer and more comfortable option for patients, encouraging regular monitoring and early detection of complications.

**Novelty**:
Current diagnostic methods for hepatic complications are often invasive or lack sensitivity and specificity. Developing non-invasive tools tailored specifically for breast cancer patients undergoing treatment is a novel approach that could revolutionize how liver health is monitored in these patients. This research fills a gap in current diagnostics by offering safer, more effective monitoring options.

| | |
|---|---|
| Significance (experts' scores): $m_S$=3.67 ($SD_S$=1.53) | |
| Novelty (experts' scores): $m_N$=4.00 ($SD_N$=0.00) | |
| Novelty (literature search): N (No, there was no same research) | |
| Search terms: (Non-Invasive Diagnostic Tools) And (Early Detection) AND (Hepatic Complications) AND (Breast Cancer Patients) | |

**3) Evaluating the Efficacy of Hepatoprotective Agents in Preventing Liver Damage During Breast Cancer Treatment**

Objective:
To assess the effectiveness of hepatoprotective agents (such as ursodeoxycholic acid and silymarin) in preventing liver damage in breast cancer patients undergoing various treatments.

**Significance**:
If hepatoprotective agents are proven effective, they could be incorporated into breast cancer treatment protocols to proactively protect the liver, thus reducing the incidence and severity of liver-related side effects. This could improve patient adherence to treatment regimens and enhance overall treatment efficacy and quality of life for breast cancer patients.

**Novelty**:
The use of hepatoprotective agents specifically for breast cancer patients undergoing treatment has not been extensively studied. Most research focuses on managing liver damage after it has occurred rather than on prevention. This project introduces a preventative approach, potentially leading to significant improvements in patient care by reducing the burden of liver-related complications.

Significance (experts' scores): $m_S$=2.33 ($SD_S$=1.53)

Novelty (experts' scores): $m_N$=3.00 ($SD_N$=1.00)

Novelty (literature search): Y (Yes, there was the same research that was already done)

Search terms: (Efficacy) AND (Hepatoprotective Agents) AND (Prevention) AND (Liver Damage) AND (Breast Cancer Treatment)

**Skin cancer**

**Topic 1.** nose_nasal_pimple_nostril

**Primary issue**: Patients with skin cancer are primarily concerned about lesions or pimples on the nose and nostrils, which may resemble common skin issues but raise fears of malignancy. They often seek clarification on whether these lesions require a biopsy or surgery, particularly after sharing photos with their clinician for evaluation. The focus is on distinguishing between benign and malignant conditions and determining the appropriate treatment plan.

**1) Comparative Study on the Accuracy of Teledermatology for Nasal Skin Lesions**

Objective:
To evaluate the diagnostic accuracy of teledermatology specifically for nasal skin lesions, comparing it to in-person evaluations by dermatologists.

**Significance**:
While teledermatology is a growing field with proven accuracy for general skin lesions, the unique anatomical and visual characteristics of nasal lesions warrant a focused study. Given the prominence of the nose in the facial structure and the potential for confusion between benign and malignant lesions, this study would provide valuable insights into the strengths and limitations of teledermatology in this specific context.

**Novelty**:
Existing research has broadly covered teledermatology for skin cancer diagnosis, but there is a

| |
|---|
| lack of studies that isolate nasal lesions. This project would fill that gap, offering new data on the effectiveness of teledermatology for this high-stakes, cosmetically significant area, potentially leading to better diagnostic protocols. |
| Significance (experts' scores): $m_S$=3.00 ($SD_S$=1.00) |
| Novelty (experts' scores): $m_N$=3.67 ($SD_N$=0.58) |
| Novelty (literature search): N (No, there was no same research) |
| Search terms: (Comparative Study) AND (Accuracy) AND (Teledermatology) AND (Nasal Skin Lesions) |
| **2) Outcomes of Non-Surgical Treatments for Pre-Cancerous Nasal Lesions**<br><br>Objective:<br>To assess and compare the effectiveness and patient satisfaction of non-surgical treatments (e.g., topical therapies, cryotherapy, laser treatments) for pre-cancerous nasal lesions against traditional surgical methods.<br><br>**Significance**:<br>Non-surgical treatments are increasingly popular due to their less invasive nature and potential for better cosmetic outcomes. This research is crucial for developing treatment guidelines that balance effectiveness with patient quality of life, particularly in visible areas like the nose where cosmetic considerations are paramount.<br><br>**Novelty**:<br>While non-surgical treatments have been studied for general skin lesions, there is limited research focused on pre-cancerous conditions specifically affecting the nasal area. This project would explore new ground by examining treatment outcomes and patient perspectives specific to nasal lesions, offering insights that could inform patient-centered care and treatment decisions. |
| Significance (experts' scores): $m_S$=2.67 ($SD_S$=1.15) |
| Novelty (experts' scores): $m_N$=4.33 ($SD_N$=0.58) |
| Novelty (literature search): N (No, there was no same research) |
| Search terms: (Non-Surgical treatments) AND (Pre-cancerous Nasal lesions) |
| **3) Psychosocial Impact of Facial Skin Cancer and Treatment on Patients**<br><br>Objective:<br>To investigate the psychological and social effects of a skin cancer diagnosis and treatment on the nose, including how different treatment modalities impact mental health, self-esteem, and social interactions.<br><br>**Significance**:<br>Skin cancer on the face, particularly the nose, can have profound psychosocial effects due to the visibility of the area. Understanding these impacts is essential for providing holistic care that addresses not only the physical but also the emotional and psychological needs of patients.<br><br>**Novelty**:<br>Although the psychosocial impacts of skin cancer have been studied, there is a notable lack of research focusing specifically on nasal lesions and the comparative impact of various treatments. This study would provide novel insights into the unique challenges faced by patients |
| with nasal skin cancer, contributing to a more comprehensive approach to their care (SpringerLink)." |
| Significance (experts' scores): $m_S$=3.00 ($SD_S$=1.00) |
| Novelty (experts' scores): $m_N$=3.67 ($SD_N$=0.58) |
| Novelty (literature search): Y (Yes, there was the same research that was already done) |
| Search terms: (Psychological Impact) AND (Facial Skin Cancer) AND (Treatment) |
| **Topic 2**. mole_moles_melanoma_concerned |
| **Primary issue**: Patients with skin cancer are primarily concerned about the potential for moles to develop into melanoma. They express anxiety over whether certain moles should be removed or further examined and seek reassurance or confirmation through appointments. The need for timely skin checks and possible mole removal is a significant focus of their concerns. |
| **1) Development and Validation of a Mobile App for Real-Time Monitoring and Assessment of Moles**<br><br>Objective:<br>To design and validate a mobile application that enables patients to photograph and monitor their moles regularly. The app will incorporate AI-based algorithms to assess changes in moles over time and provide risk assessments, with options for direct communication with dermatologists for high-risk cases.<br><br>**Significance**:<br>This project aims to empower patients with a tool that facilitates early detection of potentially malignant moles, potentially reducing delays in diagnosis and treatment. By integrating AI, the app can assist in prioritizing dermatology appointments for those at higher risk, thus improving resource allocation and patient outcomes.<br><br>**Novelty**:<br>Although several mobile apps exist for skin cancer risk assessment, this project would focus on improving the accuracy of AI algorithms and integrating real-time monitoring with direct clinician communication. The novelty lies in the seamless integration of patient self-monitoring with professional oversight, addressing existing gaps in accuracy and user engagement. |
| Significance (experts' scores): $m_S$=3.33 ($SD_S$=1.53) |
| Novelty (experts' scores): $m_N$=4.00 ($SD_N$=1.00) |
| Novelty (literature search): N (No, there was no same research) |
| Search terms: (Real-time Moles Monitoring) AND (Mobile app) AND (Validation) |
| **2) Longitudinal Study on Patient Anxiety and Decision-Making in Mole Surveillance and Removal**<br><br>Objective:<br>To investigate how anxiety influences patient decision-making regarding mole surveillance and removal, examining the psychological factors that lead to either proactive or delayed responses to mole changes. This study will track patients over time to correlate anxiety levels with clinical outcomes.<br><br>**Significance**:<br>Understanding the psychological drivers behind patients' decisions can lead to better patient education and support, potentially reducing unnecessary procedures while ensuring timely |

intervention for those at actual risk. It can also help clinicians develop more empathetic communication strategies that address patient concerns effectively.

**Novelty**:
While there is some research on patient anxiety in dermatology, this study's unique approach of longitudinally tracking anxiety-related decision-making in mole surveillance is relatively unexplored. The results could significantly enhance personalized patient care strategies in dermatology, making this a pioneering study in its domain.

| | |
|---|---|
| Significance (experts' scores): $m_S$=2.33 ($SD_S$=1.53) | |
| Novelty (experts' scores): $m_N$=2.67 ($SD_N$=1.53) | |
| Novelty (literature search): N (No, there was no same research) | |
| Search terms: (Patient Anxiety) AND (Decision-Making) AND (Mole Surveillance) AND (Removal) | |

**3) Impact of Teledermatology on Early Detection and Management of Melanoma in High-Risk Patients**

Objective:
To assess the effectiveness of teledermatology services in early detection and management of melanoma among patients identified as high-risk due to factors such as family history or the presence of atypical moles. The study would compare outcomes between teledermatology and traditional in-person consultations.

**Significance**:
This research could provide valuable insights into the potential of teledermatology to bridge gaps in care, especially for high-risk patients who may not have easy access to in-person dermatological services. By focusing on a high-risk population, the study aims to demonstrate the benefits and limitations of teledermatology in a critical subgroup.

**Novelty**:
Although teledermatology has been studied broadly, its specific impact on high-risk patients for melanoma has not been comprehensively explored. This project would fill that gap, providing new data on how teledermatology can be optimized to serve those most in need of early intervention.

| | |
|---|---|
| Significance (experts' scores): $m_S$=2.33 ($SD_S$=1.15) | |
| Novelty (experts' scores): $m_N$=3.33 ($SD_N$=0.58) | |
| Novelty (literature search): Y (Yes, there was the same research that was already done) | |
| Search terms: (Teledermatology) AND (early detection) AND (Melanoma) AND (High-Risk Patients) | |

**Topic 3**. ear_earlobe_ears_earrings

**Primary issue**: Patients with skin cancer are likely concerned about issues related to the ear and earlobe, particularly in relation to surgical interventions like biopsies and the impact on hearing. They may also be worried about complications involving the cartilage and ear canal, especially following surgeries or procedures that affect these areas. Additionally, concerns about wearing earrings and the overall appearance of the ear post-surgery are likely prevalent.

**1) Impact of Ear and Earlobe Reconstruction on Hearing Post Skin Cancer Surgery**

Objective:
To assess the long-term effects of reconstructive surgery on hearing in patients who have undergone ear or earlobe surgery for skin cancer, focusing on both functional and aesthetic outcomes.

**Significance**:
While ear reconstruction following skin cancer surgery is well-studied, the specific impact on hearing is less explored. This study could fill an important gap by addressing whether these procedures compromise auditory function, which is crucial for patient quality of life.

**Novelty**:
Although ear reconstruction techniques are extensively documented, the specific correlation between these procedures and post-operative hearing outcomes has not been comprehensively studied, making this project unique and valuable (SpringerLink) (PubMed).

| | |
|---|---|
| Significance (experts' scores): $m_S$=2.33 ($SD_S$=0.58) | |
| Novelty (experts' scores): $m_N$=3.00 ($SD_N$=1.73) | |
| Novelty (literature search): N (No, there was no same research) | |
| Search terms: (Ear reconstruction) AND (Hearing) AND (Skin Cancer Surgery) | |

**2) Investigation of Cartilage-Sparing Techniques in Skin Cancer Surgery of the Ear**

Objective:
To develop and evaluate surgical techniques that prioritize cartilage preservation during skin cancer removal from the ear, aiming to optimize both structural integrity and cosmetic outcomes.

**Significance**:
Cartilage integrity is vital for both the function and appearance of the ear. This study is significant because it would help refine surgical practices to balance cancer removal with minimal cartilage damage, improving patient outcomes.

**Novelty**:
While some research exists on cartilage-sparing methods, this study would be novel in focusing specifically on skin cancer surgeries, providing tailored strategies for this patient group (SpringerLink) (PubMed).

| | |
|---|---|
| Significance (experts' scores): $m_S$=2.67 ($SD_S$=1.15) | |
| Novelty (experts' scores): $m_N$=2.67 ($SD_N$=1.15) | |
| Novelty (literature search): Y (Yes, there was the same research that was already done) | |
| Search terms: (Cartilage-sparing techniques) AND (Skin Cancer Surgery) AND (Ear) | |

**3) Assessment of Quality of Life and Psychosocial Impact Following Ear Surgery in Skin Cancer Patients**

Objective:
To evaluate the psychosocial outcomes and quality of life in patients who have undergone ear surgery for skin cancer, with a particular focus on body image, self-esteem, and social interactions.

**Significance**:
Beyond physical recovery, the emotional and social impacts of ear surgery are critical to patient care. This study would provide valuable insights into how surgery affects patients' lives, helping

| |
|---|
| clinicians offer more holistic care. <br><br> **Novelty:** <br> While the psychological impact of cancer surgeries is well-documented, research specifically addressing the quality of life and psychosocial effects following ear surgeries for skin cancer remains limited, offering a fresh perspective in this area (BioMed Central). |
| Significance (experts' scores): $m_S$=3.00 ($SD_S$=1.00) |
| Novelty (experts' scores): $m_N$=3.33 ($SD_N$=2.08) |
| Novelty (literature search): N (No, there was no same research) |
| Search terms: (Quality of life) AND (Psychosocial impact) AND (ear surgery) AND (Skin Cancer) |
| **Topic 4**. efudix_efudex_effudex_treatment |
| **Primary issue**: Patients with skin cancer are primarily concerned with the treatment process involving Efudex (Efudix/Effudex), focusing on its application duration, typically over several weeks, and its effectiveness. They also express concerns about the treatment's impact, possibly following a biopsy, and how well it is treating their condition. These messages suggest a need for clearer guidance on treatment expectations and outcomes. |
| **1) Personalized Dosing Strategies for Efudex in Skin Cancer Treatment** <br><br> Objective: <br> To develop and validate personalized dosing regimens for Efudex based on patient-specific factors such as skin type, genetic markers, and cancer severity. <br><br> **Significance**: <br> This project aims to improve the effectiveness of Efudex treatment while minimizing adverse effects, addressing the variability in patient responses that currently exists. <br><br> **Novelty**: <br> There is a lack of research specifically focusing on personalized dosing strategies for Efudex, making this study innovative in optimizing patient outcomes through individualized care. |
| Significance (experts' scores): $m_S$=3.00 ($SD_S$=1.73) |
| Novelty (experts' scores): $m_N$=3.00 ($SD_N$=1.73) |
| Novelty (literature search): N (No, there was no same research) |
| Search terms: (Personalized Dosing) AND (Efudex) AND (Skin Cancer) |
| **2) Impact of Patient Education on Efudex Treatment Adherence and Outcomes** <br><br> Objective: <br> To assess the effectiveness of comprehensive patient education programs in improving adherence to Efudex treatment and enhancing clinical outcomes. <br><br> **Significance**: <br> Improving patient understanding and management of Efudex treatment could lead to better adherence, more effective treatment, and reduced anxiety, ultimately improving patient satisfaction. <br><br> **Novelty**: <br> While patient education is a known component of treatment, specific studies linking it to Efudex |
| treatment adherence and outcomes are scarce, making this research crucial for evidence-based patient support strategies. |
| Significance (experts' scores): $m_S$=1.67 ($SD_S$=0.58) |
| Novelty (experts' scores): $m_N$=2.33 ($SD_N$=1.53) |
| Novelty (literature search): N (No, there was no same research) |
| Search terms: (Patient Education) AND (Efudex Adherence) AND (Outcomes) |
| **3) Long-Term Effects of Efudex Treatment on Skin Quality and Cancer Recurrence Rates**<br><br>Objective:<br>To investigate the long-term effects of Efudex on skin quality, including scarring and pigmentation changes, as well as its impact on skin cancer recurrence rates.<br><br>**Significance**:<br>Understanding the long-term impact of Efudex is essential for providing patients with accurate information about the potential risks and benefits of treatment over time.<br><br>**Novelty**:<br>Current research focuses primarily on the short-term efficacy and side effects of Efudex, leaving a gap in understanding its long-term consequences, which this study aims to fill. |
| Significance (experts' scores): $m_S$=3.00 ($SD_S$=1.00) |
| Novelty (experts' scores): $m_N$=2.67 ($SD_N$=1.53) |
| Novelty (literature search): N (No, there was no same research) |
| Search terms: (Long-Term Effects) AND (Efudex) AND (Skin Quality) AND (Cancer Recurrence) |
| **Topic 5**. stitches_incision_stitch_sutures |
| **Primary issue**: Patients with skin cancer are primarily concerned about the management of their surgical wounds, specifically issues related to stitches and sutures. Their messages reflect worries about incision care, the healing process, and potential complications with stitching after surgery. They seek guidance on proper wound care and expect timely advice on how to manage their sutures effectively. |
| **1) Comparative Study on Healing Outcomes with Different Suture Techniques in Skin Cancer Surgery**<br><br>Objective:<br>To assess the efficacy of various suture techniques in skin cancer surgery by comparing healing times, complication rates, and patient satisfaction.<br><br>**Significance**:<br>The findings could standardize the best suture practices, improving surgical outcomes and patient experience.<br><br>**Novelty**:<br>While suture techniques have been studied in general surgery, their specific impact on skin cancer surgeries remains underexplored. |
| Significance (experts' scores): $m_S$=3.00 ($SD_S$=1.00) |
| Novelty (experts' scores): $m_N$=3.33 ($SD_N$=0.58) |
| Novelty (literature search): Y (Yes, there was the same research that was already done) |
| Search terms: (Outcomes) AND (Suture Techniques) AND (Skin Cancer Surgery) |

**2) Development and Evaluation of a Patient-Centered Digital Tool for Post-Surgical Wound Care**

Objective:
To create and evaluate a digital tool tailored to guide skin cancer patients in managing their post-surgical wounds.

**Significance**:
The tool could significantly enhance patient compliance, reduce complications, and alleviate anxiety related to wound care.

**Novelty**:
This project fills a gap in the market, as existing digital tools do not cater specifically to the unique needs of skin cancer patients.

Significance (experts' scores): $m_S$=2.00 ($SD_S$=0.00)

Novelty (experts' scores): $m_N$=1.67 ($SD_N$=0.58)

Novelty (literature search): N (No, there was no same research)

Search terms: (Development) AND (Evaluation) AND (Patient-Centered) AND (Digital Tool) AND (Post-Surgical Wound Care)

**3) Impact of Suture Materials on Scar Formation in Skin Cancer Patients**

Objective:
To explore how different suture materials affect scar formation in skin cancer patients, considering skin type and incision location.

**Significance**:
The research could lead to optimized suture material choices that minimize visible scarring, addressing a major patient concern.

**Novelty**:
Scar formation related to specific suture materials in skin cancer surgeries has not been comprehensively studied, making this project particularly valuable.

Significance (experts' scores): $m_S$=2.67 ($SD_S$=1.15)

Novelty (experts' scores): $m_N$=2.67 ($SD_N$=0.58)

Novelty (literature search): N (No, there was no same research)

Search terms: (Suture materials) AND (Scar Formation) AND (Skin Cancer)